\documentclass[10pt,conference]{IEEEtran}

\usepackage{booktabs}
\usepackage{multirow}
\usepackage{pgfplots}
\pgfplotsset{compat=1.18}
\usepackage[numbers,sort&compress]{natbib}
\usepackage{url}
\usepackage{amsmath}
\usepackage{xcolor}
\usepackage{subcaption}
\usepackage{tikz}
\usetikzlibrary{shapes.geometric,arrows.meta,positioning,fit,calc}

\title{Knowledge Access Beats Model Size: Memory Augmented Routing for Persistent AI Agents}

\author{
  Xunzhuo Liu$^{1}$, Bowei He$^{2,3, \dagger}$, Xue Liu$^{1,2,3,4}$, Andy Luo$^{5}$, Haichen Zhang$^{5}$, Huamin Chen$^{6}$ \\
  $^1$ vLLM Semantic Router Project, $^2$ MBZUAI, $^3$ McGill University, $^4$ Mila, $^5$ AMD, $^6$ Red Hat \\
  $\dagger$ Corresponding author: \texttt{Bowei.He@mbzuai.ac.ae}
}

\begin{document}

\maketitle

\begin{abstract}

Production AI agents frequently receive user-specific queries that are highly repetitive, with up to 47\% being semantically similar to prior interactions, yet each query is typically processed with the same computational cost.  We argue that this redundancy can be exploited through conversational memory, transforming repetition from a cost burden into an efficiency advantage. We propose a memory-augmented inference framework in which a lightweight 8B-parameter model leverages retrieved conversational context to answer all queries via a low-cost inference path. Without any additional training or labeled data, this approach achieves 30.5\% F1, recovering 69\% of the performance of a full-context 235B model while reducing effective cost by 96\%. Notably, a 235B model without memory (13.7\% F1) underperforms even the standalone 8B model (15.4\% F1), indicating that for user-specific queries, access to relevant knowledge outweighs model scale. We further analyze the role of routing and confidence. At practical confidence thresholds, routing alone already directs 96\% of queries to the small model, but yields poor accuracy (13.0\% F1) due to confident hallucinations. Memory does not substantially alter routing decisions; instead, it improves correctness by grounding responses in retrieved user-specific information. As conversational memory accumulates over time, coverage of recurring topics increases, further narrowing the performance gap. We evaluate on 152 LoCoMo questions (Qwen3-8B/235B) and 500 LongMemEval questions. Incorporating hybrid retrieval (BM25 + cosine similarity) improves performance by an additional +7.7 F1, demonstrating that retrieval quality directly enhances end-to-end system performance. Overall, our results highlight that memory, rather than model size, is the primary driver of accuracy and efficiency in persistent AI agents.
\end{abstract}

\section{Introduction}
\label{sec:intro}

Large language model inference costs scale with both model size and input length.
Production systems that answer user-specific questions, such as personal assistants, customer support agents, multi-turn dialogue, mostly face a cost--quality tension: large models (100B+ parameters) produce better answers but cost 10--30$\times$ more per token than small models (7--8B).
Meanwhile, small models lack access to knowledge accumulated across prior conversations.
Without that knowledge, a small model asked ``What is Caroline's marital status?'' can only guess, because it has never seen the conversation where Caroline mentioned being single.

\paragraph{Agent workloads are repetitive.}
This cost--quality tension is amplified by a structural property of real agent workloads: \emph{most queries recur}.
Analysis of 100K+ production LLM traces shows that 47\% of queries are semantically similar to a previous query and 18\% are exact duplicates, while only 35\% are genuinely novel~\citep{novalogiq2026cache}.
In customer support, 40\% of tickets are repetitive~\citep{relay2018repetitive}.
In IoT personal assistants, approximately 30\% of user requests are identical or semantically similar, enabling 93\% plan reuse~\citep{agentreuse2024}.
Open-source agent frameworks such as OpenClaw~\citep{openclaw2025}, which supports 24/7 personal assistants across messaging platforms with persistent memory and hybrid search, exemplify the deployment target: a long-lived agent that repeatedly answers user-specific questions across sessions, building a growing memory store over time.

In this setting, the first query about a topic may require the expensive large model.
But every subsequent similar query can draw on memories from that first interaction, enabling the cheap small model to answer confidently.
The more repetitive the workload, the greater the amortized cost saving.
This is the production scenario our compound strategy targets.

Two lines of research address different sides of this problem.
Retrieval-augmented generation (RAG)~\citep{lewis2020rag} and its conversational variants (Mem0~\citep{mem0}, MemGPT~\citep{packer2024memgpt}) store and retrieve facts from prior interactions, improving answer quality.
Model routing~\citep{chen2023frugalgpt,ong2024routellm} directs easy queries to small models and hard queries to large ones, reducing cost.

\paragraph{The gap.}
These two techniques have been studied \emph{only in isolation}.
Memory work assumes a fixed model and measures answer quality; routing work assumes a fixed input and measures cost.
No prior work has investigated how they interact when deployed together.
This gap matters because the interaction is non-obvious.
Memory injection could \emph{help} routing by making the small model more confident about user-specific facts, reducing escalation to the expensive large model.
Or it could \emph{hurt} by injecting irrelevant context that confuses the model's confidence signal, triggering unnecessary escalation~\citep{shi2023large}.
Without empirical study, practitioners building production systems have no guidance on whether to combine these components or treat them as independent optimizations.

\paragraph{Key finding: knowledge access beats model size.}
The 235B model \emph{without} memory (13.7\% F1) is worse than the 8B (15.4\%) on user-specific questions, indicating that model size cannot substitute for missing knowledge.
A 2$\times$2 factorial (memory $\times$ routing) isolates the mechanism:

\begin{center}
\scriptsize
\begin{tabular}{lcc}
\toprule
& No routing & Routing ($\tau\!=\!0.50$) \\
\midrule
No memory & 15.4\% F1 & 13.0\% F1 \textit{(96\% on 8B)} \\
Memory    & 30.1\% F1 & \textbf{30.5\% F1} \textit{(100\% on 8B)} \\
\bottomrule
\end{tabular}
\end{center}

\noindent At a practical confidence threshold, routing \emph{already} keeps nearly all queries on the small model even without memory.
The problem is that without memory, the small model is \textbf{confidently wrong}: it fabricates plausible-sounding answers about user-specific facts it has never seen (13.0\% F1).
Memory does not primarily change the routing decision; instead, it changes whether the routed answer is \emph{correct}.
The compound effect is not ``memory enables routing'' but ``\textbf{memory makes routing worthwhile}'': routing provides cost savings, memory ensures those savings come with quality.

Consider a concrete example: a user asks ``What is Caroline's marital status?''
Without memory, the 8B typically produces a plausible but incorrect response or occasionally hedges (confidence $c = 0.38$, below $\tau = 0.50$).
With the relevant turn-pair retrieved (``Q: Are you seeing anyone? A: No, I'm single''), the 8B answers correctly with $c = 0.93$.
In aggregate, 96\% of cold queries pass the confidence threshold regardless, but their answers are wrong.
Memory transforms confidently wrong into confidently right.

\paragraph{Why this matters for production.}
The compound strategy requires no training, no labeled data, and no external classifiers, relying only the model's own log-probabilities for the confidence signal.
Memories accumulated from \emph{any} model interaction, including expensive large-model responses, become retrieval context for the cheap small model, amortizing past costs across all future queries.
In the repetitive agent workloads described above, this creates a virtuous cycle: the first query on a topic builds memory; when the same or a similar question recurs as 65\% of production queries do, the small model retrieves the prior knowledge and answers correctly instead of confidently fabricating.
Over time, the memory store grows to cover the user's recurring needs, and the quality of routed answers converges toward large-model quality at small-model cost.
A single routing layer implements the full pipeline: retrieve, inject, probe, route.

\paragraph{Contributions.}
\begin{enumerate}
\item \textbf{Identifying an unstudied interaction.} We conduct a 2$\times$2 factorial experiment (memory $\times$ routing) revealing a previously unreported phenomenon: at practical confidence thresholds, routing keeps nearly all queries on the small model \emph{regardless} of memory; but without memory, the model is \textbf{confidently wrong}. Memory does not enable routing; it makes routing worthwhile. No prior work has studied or quantified this interaction.
\item \textbf{Quantifying the production benefit.} On the full 152-question LoCoMo~\citep{locomo2024} benchmark, the compound strategy achieves +15.0~F1 over the unaided small model while keeping 100\% of queries on the cheap path, which brings a 96\% cost reduction vs.\ the large model, with zero additional training.
\item \textbf{Characterizing failure modes.} Turn-pair memories are most effective for factual recall (+28.9~F1 for single-hop) and counterproductive for temporal reasoning ($-3.8$~F1), identifying where structured memory representations are needed.
\item \textbf{Strengthening the synergy with hybrid retrieval.} On LongMemEval~\citep{longmemeval2025} (500 questions, $\sim$48 sessions per question), hybrid retrieval (BM25 + cosine) improves QA by +7.7~F1 over cosine-only search, demonstrating that retrieval quality flows through the full pipeline: better memories $\to$ higher confidence $\to$ more routing to the cheap path.
\end{enumerate}

\section{Related Work}
\label{sec:related}

\paragraph{Conversational memory.}
Mem0~\citep{mem0} uses graph-based memory extraction with entity resolution, achieving state-of-the-art on LoCoMo with a GPT-4o backbone.
MemGPT~\citep{packer2024memgpt} proposes OS-inspired hierarchical memory with paging for unbounded context.
Zep and LangMem provide memory as application SDKs.
These systems study memory in isolation.
We study how memory \emph{interacts with routing}: does retrieved context help or hurt the small model's confidence signal?

\paragraph{Model routing and cascading.}
FrugalGPT~\citep{chen2023frugalgpt} cascades models from cheap to expensive, using a learned scoring function on the model's output logprobs to decide whether to accept or escalate.
RouteLLM~\citep{ong2024routellm} trains a preference classifier for routing decisions.
AutoMix~\citep{automix2024} uses few-shot self-verification for escalation.
Dekoninck et~al.~\citep{dekoninck2025cascaderouting} unify routing and cascading into a single optimal framework, proving that the quality of the confidence estimator is the critical factor.
These methods treat routing as an independent optimization.
We show that routing \emph{composes} with memory injection, and that the ``quality estimator'' problem identified by Dekoninck et~al.\ is entangled with memory: without memory, the confidence signal is high but the answers are wrong.

\paragraph{Logprob-based confidence estimation.}
Using token-level log-probabilities as a confidence signal for LLMs has a growing empirical foundation.
Kadavath et~al.~\citep{kadavath2022language} show that language models are reasonably well-calibrated when evaluating their own outputs, with larger models better “knowing what they know,” establishing that self-assessed confidence is a viable signal.
Guo et~al.~\citep{guo2017calibration} demonstrate that modern neural networks are often miscalibrated, but that simple post-hoc corrections (temperature scaling) can restore calibration; our normalization floor $\ell_{\min}$ serves a similar role.
Varshney et~al.~\citep{varshney2023stitch} use low token-level log-probabilities to detect hallucination-prone tokens in real time, reducing hallucination rates from 47.5\% to 14.5\%.
Our routing mechanism operationalizes the same signal for a different purpose: rather than flagging individual tokens, we aggregate mean log-probability across the full response to make a binary route-or-escalate decision.
Xiong et~al.~\citep{xiong2024llmuncertainty} systematically compare white-box (logprob) and black-box (verbalized) confidence elicitation, finding that white-box methods outperform verbalized confidence and that LLMs are systematically overconfident when asked to verbalize uncertainty, motivating our use of raw logprobs rather than prompting the model to self-assess.
Our 2$\times$2 factorial adds a new dimension to this literature: the confidence signal is \emph{decoupled from correctness} for user-specific questions, because the model can be confidently wrong about facts absent from its training data.

\paragraph{Knowledge transfer across model sizes.}
Knowledge distillation~\citep{hinton2015distilling} transfers knowledge by training a student to mimic a teacher's outputs, permanently modifying the student's weights.
RetriKT~\citep{liu2023retrikt} introduces a retrieval-based alternative: knowledge from a large model is extracted into a store that a small model queries at inference time, decoupling knowledge from model parameters.
Our cross-model memory injection extends this principle to conversational systems: turn-pair memories accumulated from all model interactions, including the large model, are available as retrieval context when subsequent queries route to the small model, without any retraining.

\paragraph{RAG noise and retrieval quality.}
Shi et~al.~\citep{shi2023large} show that irrelevant retrieved context degrades LLM performance.
This motivates our design choice to store verbatim conversation turn-pairs rather than LLM-generated summaries, avoiding the hallucination risk that would compound with retrieval noise.

\paragraph{Hybrid retrieval for RAG.}
Dense retrieval~\citep{karpukhin2020dense} embeds queries and documents into a shared vector space, excelling at semantic matching but missing lexical cues.
BM25~\citep{robertson2009bm25} captures exact keyword overlap but lacks semantic generalization.
Hybrid approaches that combine both via reciprocal rank fusion~\citep{cormack2009reciprocal} or learned score combination consistently outperform either alone on open-domain QA benchmarks~\citep{ma2024finetuning}.
We study hybrid retrieval in the conversational memory setting, where memories are stored as turn-pairs and queries range from factual lookups (where BM25 excels) to semantic reasoning (where dense retrieval is essential).

\paragraph{Repetition in agent workloads.}
Production LLM agents exhibit high query redundancy: analysis of 100K+ traces shows 47\% of queries are semantically similar to a previous one~\citep{novalogiq2026cache}, and 40\% of customer support tickets are repetitive~\citep{relay2018repetitive}.
Existing approaches exploit this through semantic caching~\citep{novalogiq2026cache} or plan reuse~\citep{agentreuse2024}, achieving up to 93\% reuse rates.
These methods cache \emph{outputs}, returning a previous response verbatim when the new query is similar enough.
Our approach is complementary: we cache \emph{knowledge} as conversational memories, which the small model uses to generate a fresh, contextually appropriate response.
This is critical for personalization agents such as OpenClaw~\citep{openclaw2025}, where user context evolves across sessions and verbatim cache hits are insufficient; the agent must reason over accumulated knowledge, not replay prior answers.

\section{Method}
\label{sec:method}

\begin{figure}[t]
\centering
\resizebox{\columnwidth}{!}{%
\begin{tikzpicture}[
  node distance=0.4cm and 0.5cm,
  box/.style={rectangle, draw, rounded corners=2pt, font=\scriptsize,
              minimum height=0.6cm, minimum width=1.5cm, align=center},
  decision/.style={diamond, draw, aspect=2.2, font=\scriptsize,
                   inner sep=1pt, align=center},
]

\node[box, fill=blue!10] (query) {User\\[-1pt]Query};
\node[box, fill=yellow!15, right=of query] (vecdb) {Memory\\[-1pt]Store};
\node[box, fill=blue!10, right=of vecdb] (augment) {Augment\\[-1pt]Prompt};
\node[box, fill=green!15, right=of augment] (small) {Small\\[-1pt]Model (8B)};
\node[decision, fill=orange!15, right=0.6cm of small] (conf) {$c\!\geq\!\tau$?};
\node[box, fill=red!15, above=0.5cm of conf] (large) {Large\\[-1pt]Model (235B)};
\node[box, fill=blue!10, right=0.6cm of conf] (resp) {Response};

\draw[-{Stealth[length=4pt]}, thick]
  (query) -- node[above, font=\tiny] {hybrid} (vecdb);
\draw[-{Stealth[length=4pt]}, thick]
  (vecdb) -- node[above, font=\tiny] {top-$k$} (augment);
\draw[-{Stealth[length=4pt]}, thick] (augment) -- (small);
\draw[-{Stealth[length=4pt]}, thick]
  (small) -- node[above, font=\tiny] {probe} (conf);
\draw[-{Stealth[length=4pt]}, thick]
  (conf) -- node[right, font=\tiny] {no} (large);
\draw[-{Stealth[length=4pt]}, thick]
  (conf) -- node[above, font=\tiny] {yes} (resp);
\draw[-{Stealth[length=4pt]}, thick] (large) -| (resp);

\draw[-{Stealth[length=4pt]}, thick, dashed]
  (resp.south) -- ++(0,-0.4) -|
  node[below, pos=0.25, font=\tiny] {store turn-pair} (vecdb.south);

\end{tikzpicture}%
}
\caption{Compound pipeline. Solid arrows: inference path; dashed arrow: cross-model memory accumulation. When $c \geq \tau$, the small model's response is accepted (cheap path); otherwise the query escalates. Both paths store the turn-pair for future retrieval.}
\label{fig:pipeline}
\end{figure}
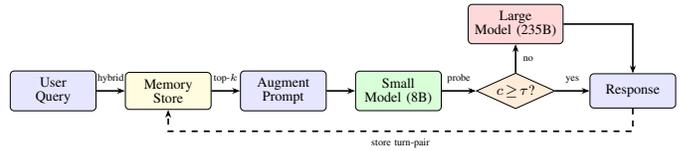

Our compound strategy combines three components in a routing layer that sits between clients and inference backends (Figure~\ref{fig:pipeline}).
We describe each component, then their interaction.

\subsection{Cross-Model Memory Injection}
After each inference call, the routing layer stores the conversation turn-pair, i.e., the user's question and the model's response, in a vector database, partitioned by user.
Each record includes:
(i)~the turn-pair text with a session timestamp prefix (e.g., ``[8~May~2023] Q: \ldots / A: \ldots''),
(ii)~an embedding from a Matryoshka representation model~\citep{kusupati2022matryoshka} at full transformer depth, and
(iii)~user-level metadata for multi-tenant isolation.

At query time, the routing layer retrieves the top-$k$ most relevant memories via hybrid search (Section~\ref{sec:method-hybrid}), combining dense cosine similarity with sparse BM25 keyword matching.
Retrieved memories are injected as system-context messages, giving the target model access to conversation history it never directly observed.

This is \emph{cross-model} because memories accumulated from interactions with any model (including the large one) are available to all subsequent models (including the small one).
The small model thus benefits from knowledge originally elicited by the large model's superior reasoning.

In a persistent agent deployment, the memory store grows over the agent's lifetime.
When a query escalates to the large model, the resulting turn-pair is stored; subsequent similar queries can draw on that memory, enabling the small model to answer confidently without re-escalation.
This creates a natural \emph{amortization} dynamic: early interactions may require the expensive model, but the cost of those escalations is amortized across all future queries that reuse the resulting memories.

\subsection{Confidence-Based Routing}

The routing layer implements a \emph{probe-then-escalate} strategy over an ordered set of models $\{M_1, M_2, \ldots\}$ sorted by increasing cost:

\begin{enumerate}
\item Send the memory-augmented prompt to the cheapest model $M_1$ with log-probability output enabled.
\item Compute the mean log-probability over all $N$ output tokens:
$\bar{\ell} = \frac{1}{N}\sum_{i=1}^{N} \log p(t_i)$.
\item Normalize to a $[0,1]$ confidence score:
\begin{equation}
\label{eq:confidence}
c = \frac{\bar{\ell} - \ell_{\min}}{|\ell_{\min}|},
\end{equation}
where $\ell_{\min}$ is a floor parameter (we use $\ell_{\min} = -3$, so $\bar{\ell} = -3$ maps to $c = 0$ and $\bar{\ell} = 0$ maps to $c = 1$).
Values below $\ell_{\min}$ are clamped to $c = 0$.
\item If $c \geq \tau$, accept $M_1$'s response. Otherwise, escalate to $M_2$ with the same augmented prompt.
\end{enumerate}

The confidence signal is \emph{zero-shot}: it requires no training data, labeled preferences, or external classifiers, relying only the model's own token-level probabilities~\citep{kadavath2022language}.
Mean sequence log-probability is a standard quality signal in LLM cascading~\citep{chen2023frugalgpt} and has been shown to correlate with factual correctness~\citep{varshney2023stitch}, though it can be miscalibrated~\citep{guo2017calibration,xiong2024llmuncertainty}.
Our normalization (Eq.~\ref{eq:confidence}) maps the raw signal to $[0,1]$ with a floor at $\ell_{\min}$, serving a role analogous to temperature scaling~\citep{guo2017calibration}.
At $\tau = 0.50$, the threshold corresponds to $\bar{\ell} = -1.5$ nats, admitting any response where the model assigns $\geq$22\% geometric-mean probability to its tokens.
The threshold $\tau$ controls the cost--quality trade-off; the normalization floor $\ell_{\min}$ controls the sensitivity range.

\subsection{The Compound Effect}

The two techniques address \emph{orthogonal} failure modes.
Routing alone keeps queries cheap but quality-blind: at practical confidence thresholds, the small model is confident about user-specific questions regardless of whether it possesses relevant knowledge. It fabricates plausible-sounding answers rather than hedging.
Memory alone provides factual grounding but does not reduce cost: without a routing mechanism, every query goes to a single model at its fixed per-token rate.

The compound effect is that routing provides cost savings and memory provides quality, and \emph{together} they achieve what neither can alone.
Formally, let $q_\text{mem}$ and $q_\text{no-mem}$ denote the small model's answer quality with and without memory.
The compound strategy is \emph{synergistic} when routing keeps the small model as the default path \emph{and} $q_\text{mem} \gg q_\text{no-mem}$: memory transforms confidently wrong answers into confidently right ones, without changing the routing decision.

In a persistent agent, this synergy has a temporal dimension.
As the memory store grows to cover the user's recurring topics, the fraction of queries that the small model can answer \emph{correctly} increases, and the quality of routed answers improves.
Given the high query repetition rates observed in production agent workloads (Section~\ref{sec:intro}), the compound strategy becomes more effective over time. The agent ``learns'' to handle familiar topics correctly \emph{and} cheaply.

This synergy is not guaranteed.
It requires two conditions: (1)~the retrieved memories must be \emph{relevant and faithful}. Irrelevant or hallucinated context would lower confidence rather than raise it (a form of RAG poisoning~\citep{shi2023large}); and (2)~the confidence signal must be \emph{calibrated}. The model's log-probabilities must actually increase when it receives useful context.
If either condition fails, the compound effect becomes sub-additive or even negative.
Our experiments test whether these conditions hold in practice.

\subsection{Hybrid Retrieval for Conversational Memory}
\label{sec:method-hybrid}

The compound strategy above retrieves memories via cosine similarity over dense embeddings.
Dense retrieval excels at semantic matching: ``Are you seeing anyone?" retrieves against ``What is Caroline's marital status?” but fails when the connection is primarily lexical.: a question about a specific restaurant name, a date, or a technical term may not share semantic neighborhood with the memory that mentions it verbatim.

Hybrid retrieval combines dense cosine search with sparse keyword matching (BM25) to capture both semantic and lexical signals.
For each query, the system executes two parallel searches against the memory store:

\begin{enumerate}
\item \textbf{Dense search.} The query is embedded with the same Matryoshka model used for memory storage. Top-$k$ candidates are retrieved by cosine similarity.
\item \textbf{Sparse search.} The query is tokenized and matched against memory text via BM25~\citep{robertson2009bm25} with n-gram expansion, capturing exact and near-exact keyword overlap.
\end{enumerate}

The two ranked lists are combined via a configurable fusion strategy.
In our implementation, the routing layer supports reciprocal rank fusion, weighted score combination, and BM25-dominant modes where sparse matches are promoted when keyword overlap is high.

This is particularly important for conversational memory because turn-pair memories contain both the user's original phrasing \emph{and} the assistant's response.
A question like ``What did I say about the Amalfi Coast trip?'' contains a named entity (``Amalfi Coast'') that BM25 can match exactly, while a semantic-only search might retrieve any travel-related memory.

\section{Experimental Setup}
\label{sec:setup}

\subsection{Models and Infrastructure}

All experiments use a single AMD Instinct MI300X GPU (192\,GB HBM3) running two models simultaneously:
\begin{itemize}
\item \textbf{Qwen3-VL-8B-Instruct} (8B parameters, BF16, 32K context).
\item \textbf{Qwen3-VL-235B-A22B-Instruct-AWQ} (235B total / 22B active MoE, AWQ quantized, 16K context).
\end{itemize}
Memory storage uses Milvus~\citep{milvus2021} with 768-dimensional embeddings from a 2D Matryoshka model~\citep{kusupati2022matryoshka} at full transformer depth.

\subsection{Datasets}

\paragraph{LoCoMo.}
We use LoCoMo~\citep{locomo2024}, a conversational memory QA benchmark, for the main memory--routing synergy experiments.
Conversation~26 contains 19 sessions between two speakers spanning 214 turns, with 152 QA questions in four categories:
single-hop (70), requiring one fact;
multi-hop (40), requiring multiple facts;
open-domain (12), requiring general knowledge alongside conversation context;
and temporal (30), requiring reasoning about when events occurred.

\paragraph{LongMemEval.}
For the hybrid retrieval ablation, we use LongMemEval~\citep{longmemeval2025}, a multi-session conversational benchmark from ICLR~2025.
Each of the 500 questions is paired with a ``haystack'' of $\sim$48 chat sessions ($\sim$550 turns, $\sim$121K tokens) in which the answer is buried.
Questions span six categories: single-session-user (70), single-session-assistant (56), single-session-preference (30), multi-session (133), temporal-reasoning (133), and knowledge-update (78).
Ground-truth \texttt{answer\_session\_ids} enable retrieval evaluation.
The scale and diversity of LongMemEval (500 distinct user histories vs.\ LoCoMo's single conversation) provides a more robust test of retrieval strategies.

\subsection{Memory Construction}

All 214 conversational turns are stored as turn-pairs: \texttt{[timestamp] Q: $\langle$user$\rangle$ / A: $\langle$assistant$\rangle$}.
Each pair is embedded and inserted into the vector database.
This simulates the steady-state of a persistent agent that has accumulated memories over its deployment lifetime, the scenario where the amortization benefit is fully realized.
In practice, memory builds incrementally: early queries may escalate more frequently, with the escalation rate decreasing as the store grows to cover the user's recurring topics.

\subsection{Evaluation Conditions}

\paragraph{LoCoMo: Memory--routing synergy (6 conditions).}
The first four conditions form a 2$\times$2 factorial design (memory $\times$ routing):
\begin{enumerate}
\item \textbf{Cold 8B}: Direct 8B, no memory, no routing. Small-model baseline.
\item \textbf{Cold compound}: No memory + confidence routing ($\tau = 0.50$). Isolates routing's effect without memory.
\item \textbf{Warm (memory only)}: Memory retrieval to 8B, no routing. Isolates memory injection.
\item \textbf{Warm (compound)}: Memory retrieval + confidence routing ($\tau = 0.50$). The full pipeline: retrieve $\to$ inject $\to$ probe 8B $\to$ escalate if needed.
\item \textbf{Cold 235B}: Direct 235B, no memory, no routing. Large-model baseline.
\item \textbf{Full-context 235B}: All 19 sessions as conversation history to 235B. Quality upper bound.
\end{enumerate}

\paragraph{LongMemEval: Hybrid retrieval evaluation.}
We compare cosine-only (dense) retrieval against hybrid retrieval (dense + BM25), both using the full routing pipeline with the 8B model.
For each of the 500 questions, all $\sim$245 turn-pair memories from the question's haystack sessions are pre-loaded into Milvus with precomputed embeddings.
A stratified sample of 100 questions (proportional to question type) is evaluated through the full routing pipeline.

\subsection{Metrics}

\textbf{F1 Score}: Token-level F1 between predicted and reference answers, following LoCoMo protocol.
\textbf{BLEU-1}: Unigram precision for answer overlap.
\textbf{Effective Cost}: $\text{EffCost} = (\text{input} + 4 \times \text{output}) \times (P / 8)$, where $P$ is model parameters in billions.
The $4\times$ output multiplier reflects the well-documented asymmetry between prefill and decode: input tokens are processed in a single parallelized forward pass, while each output token requires a separate memory-bandwidth-bound pass through the full model weights~\citep{agrawal2024sarathi}.
This matches commercial pricing, e.g., OpenAI charges 4$\times$ more per output token than per input token across GPT-4o and GPT-4.1.
\textbf{Routing distribution}: Fraction of queries served by the small model vs.\ escalated.

\section{Results}
\label{sec:results}

\subsection{Memory Injection Recovers Large-Model Quality at Small-Model Cost}

\begin{table}[t]
\centering
\scriptsize
\caption{F1 scores (\%) on LoCoMo by category (152 questions). S = single-hop (70), M = multi-hop (40), O = open (12), T = temporal (30). Cold compound = routing without memory ($\tau\!=\!0.50$, 96\% on 8B). ``---'' = unreported in source.}
\label{tab:f1}
\begin{tabular}{lccccc}
\toprule
Condition & S & M & O & T & ALL \\
\midrule
Warm (compound)    & 46.4 & 20.8 & 25.5 & 10.3 & \textbf{30.5} \\
Warm (memory only) & 45.7 & 20.8 & 25.5 & 10.3 & 30.1 \\
Cold compound      & 16.7 &  6.2 &  9.6 & 13.1 & 13.0 \\
Cold 8B            & 17.5 &  8.8 & 24.7 & 14.1 & 15.4 \\
Cold 235B          & 19.8 &  7.6 & 16.1 &  6.4 & 13.7 \\
Full-ctx 235B      & 61.9 & 47.0 & 19.1 & 16.1 & 43.9 \\
\midrule
\multicolumn{6}{l}{\textit{Published baselines (Mem0 / LoCoMo):}} \\
Mem0 (GPT-4o)      & --- & 38.7 & 47.7 & 48.9 & 41.0 \\
Zep                & --- & 35.7 & 49.6 & 42.0 & 36.7 \\
LangMem            & --- & 35.5 & 40.9 & 30.8 & 33.3 \\
\bottomrule
\end{tabular}
\end{table}

Table~\ref{tab:f1} shows the main result.
Memory injection raises the 8B model from 15.4\% to 30.5\% F1, a \textbf{+15.0 point improvement}, recovering \textbf{69\%} of the full-context 235B quality (43.9\%).
The compound condition (Warm) achieves nearly identical F1 to memory-only (30.5 vs.\ 30.1), confirming that routing does not sacrifice quality.

Model size alone does not help: cold 235B (13.7\% F1) is \emph{worse} than cold 8B (15.4\%), because neither possesses user-specific knowledge and the 235B's verbose hedging further dilutes token-level F1.
Memory access, not parameter count, determines quality on personalization tasks.

The memory-augmented 8B (30.5\%) approaches Mem0 with GPT-4o (41.0\%) using a model roughly 50$\times$ smaller.
The remaining gap reflects Mem0's entity resolution, structured memory graphs, and GPT-4o's stronger reasoning, not the memory injection principle itself.

\subsection{The 2$\times$2 Factorial: Memory Makes Routing Worthwhile}

\begin{table}[t]
\centering
\small
\caption{2$\times$2 factorial: memory $\times$ routing (152 questions, Qwen3-8B, $\tau\!=\!0.50$). ``\% on 8B'' = fraction of queries served by the small model. Memory does not change routing behavior; instead it changes answer \emph{quality}.}
\label{tab:factorial}
\begin{tabular}{lccc}
\toprule
& F1 (\%) & \% on 8B & EffCost \\
\midrule
\multicolumn{4}{l}{\textit{No memory}} \\
\quad No routing (Cold 8B) & 15.4 & 100 & 15K \\
\quad Routing (Cold compound) & 13.0 & 96.1 & $\sim$15K \\
\midrule
\multicolumn{4}{l}{\textit{With memory}} \\
\quad No routing (Warm mem-only) & 30.1 & 100 & 110K \\
\quad Routing (Warm compound) & \textbf{30.5} & \textbf{100} & \textbf{22K} \\
\bottomrule
\end{tabular}
\end{table}

Routing behavior is nearly independent of memory (Table~\ref{tab:factorial}).
At $\tau = 0.50$, 96\% of queries stay on the 8B model \emph{without} memory and 100\% stay \emph{with} memory.
The difference is answer quality: cold compound achieves 13.0\% F1 while warm compound achieves 30.5\%.
The small model without memory does not hedge. It fabricates plausible answers about user-specific facts it has never seen, with sufficient token-level confidence to pass the routing threshold.

Memory and routing therefore address orthogonal failure modes.
Routing provides cost savings (nearly all queries on the cheap path regardless of memory).
Memory provides correctness ($+$17.5~F1 over cold compound).
The compound strategy is the only condition that achieves both: 30.5\% F1 at 22K EffCost.

Per-category cold compound routing confirms this pattern across question types: single-hop (100\% on 8B, 16.7\% F1), multi-hop (94\% on 8B, 6.2\% F1), open-domain (100\% on 8B, 9.6\% F1), temporal (89\% on 8B, 13.1\% F1).
The 4\% of queries that do escalate are multi-hop and temporal questions where uncertainty occasionally surfaces.

\subsection{Cost Analysis: The Compound Strategy Is Pareto Optimal}

\begin{table}[b]
\centering
\scriptsize
\caption{Token usage and effective cost (152 questions). EffCost = $(\text{in} + 4 \times \text{out}) \times P/\text{8B}$, where the $4\times$ output multiplier reflects costlier autoregressive decode and $P$ is model parameters. All Warm conditions run on 8B ($P/\text{8B}=1$); 235B conditions cost $\approx$29$\times$ more per token.}
\label{tab:cost}
\begin{tabular}{lrrr}
\toprule
Condition & In & Out & EffCost \\
\midrule
Cold 8B          &      9.6K & 1.4K &       15K \\
Cold compound    &     $\sim$10K & $\sim$1.4K & $\sim$16K \\
Cold 235B        &      9.6K & 1.4K &      443K \\
Warm (mem.\ only) &    103K & 1.7K &      110K \\
Warm (compound)  &     16K & 1.5K &       22K \\
Full-ctx 235B    &  2,314K & 1.8K &   68,179K \\
\bottomrule
\end{tabular}
\end{table}

Cost is dominated by model size, not token count (Table~\ref{tab:cost}).
Cold 8B and Cold 235B process identical tokens (9.6K input, 1.4K output), yet the 235B costs 29$\times$ more because EffCost scales linearly with model parameters.

Memory injection adds input tokens but the compound pipeline mitigates this.
Warm (memory only) consumes 102K input tokens, 10$\times$ the cold baseline, from injected turn-pair memories.
Warm (compound) consumes only 16K because the confidence routing probe uses a truncated context window.
Both Warm conditions achieve nearly identical F1 (30.1 vs.\ 30.5), confirming the probe captures sufficient context.

The compound strategy is Pareto optimal: 30.5\% F1 at 22K EffCost, \textbf{96\% cheaper} than cold 235B (443K) and \textbf{99.97\% cheaper} than full-context 235B (68M), at 2.2$\times$ the F1 of cold 235B.
This optimality arises because the compound strategy achieves cheap routing (100\% on 8B) \emph{and} the quality that makes cheap routing worthwhile. Routing alone achieves the former (Table~\ref{tab:factorial}), but not the latter.

\subsection{Where Memory Helps and Where It Does Not}
\begin{figure}[t]
    \centering
    \includegraphics[width=0.9\columnwidth]{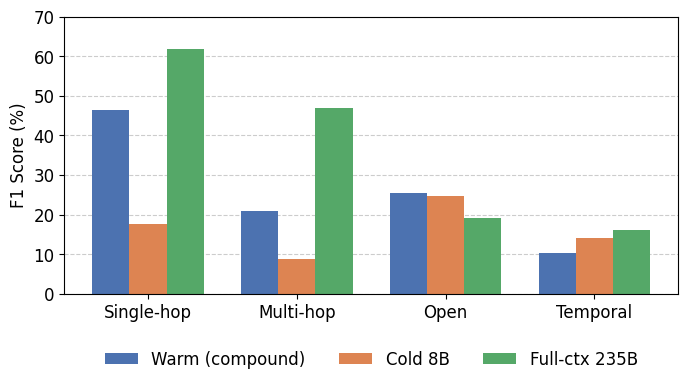}
    \caption{F1 by question category. Memory injection provides the largest gains 
    for factual-recall categories (single-hop, multi-hop) and the smallest for 
    categories requiring external knowledge or temporal reasoning.}
    \label{fig:category}
\end{figure}

Memory injection helps most when the answer exists verbatim in a retrievable turn, and least when it does not (Figure~\ref{fig:category}).

\textbf{Single-hop} questions show the largest gain: +28.9~F1 (46.4 vs.\ 17.5).
These require one specific fact (``What is Caroline's job?''), and the retrieved turn-pair directly contains the answer.

\textbf{Multi-hop} questions also benefit substantially (+12.0~F1).
The top-$k$ retrieved memories can supply some of the required facts, though combining them remains challenging for the 8B model.

\textbf{Open-domain} questions show minimal change (+0.8~F1): these require world knowledge that conversational memories do not contain.

\textbf{Temporal} questions \emph{decline} ($-3.8$~F1).
Despite timestamp prefixes in stored memories, the model struggles to reason about ``when'' from unstructured date text.
This is the clearest signal that turn-pair memories are insufficient for temporal reasoning; structured temporal indices or date-range filtering may be needed.

\subsection{Hybrid Retrieval Improves Memory Quality at Scale}
\label{sec:hybrid-ablation}

The LoCoMo experiments above use cosine-only retrieval from a single conversation (19~sessions, 214~turns).
To test retrieval strategies at realistic scale, we evaluate on LongMemEval~\citep{longmemeval2025}: 500 questions, each with $\sim$48 haystack sessions ($\sim$245 turn-pair memories, $\sim$121K tokens).
We compare both retrieval strategies through the full routing pipeline with the 8B model.

\begin{table}[t]
\centering
\small
\caption{End-to-end QA on LongMemEval (100 stratified questions, 8B model). Hybrid = BM25 + cosine fusion. Best per column in bold.}
\label{tab:hybrid}
\begin{tabular}{lcc}
\toprule
Retrieval & F1 (\%) & BLEU-1 (\%) \\
\midrule
Cosine-only                         & 36.5 & 34.9 \\
\textbf{Hybrid (BM25 + cosine)}     & \textbf{44.2} & \textbf{42.0} \\
\bottomrule
\end{tabular}
\end{table}

\paragraph{Findings (Table~\ref{tab:hybrid}).}
Hybrid retrieval yields \textbf{+7.7~F1} (36.5\% $\to$ 44.2\%), a 21\% relative improvement over cosine-only search.
The gain is not uniform across question types. It concentrates where keyword overlap between questions and memories is high, revealing a complementarity between dense and sparse retrieval that we analyze next.

\begin{table}[t]
\centering
\small
\caption{Per-type F1 (\%) on LongMemEval: cosine-only vs.\ hybrid retrieval. $\Delta$ = absolute F1 gain from adding BM25.}
\label{tab:hybrid-pertype}
\begin{tabular}{lrrrc}
\toprule
Question type & $n$ & Cosine & Hybrid & $\Delta$ \\
\midrule
single-session-user       & 14 & 57.9 & \textbf{76.9} & +19.0 \\
single-session-assistant  & 11 & 64.8 & 61.8          & $-$3.0 \\
single-session-preference &  6 &  2.7 &  5.8          & +3.1 \\
multi-session             & 27 & 29.0 & 26.8          & $-$2.2 \\
temporal-reasoning        & 26 & 28.7 & \textbf{34.4} & +5.7 \\
knowledge-update          & 16 & 36.3 & \textbf{63.0} & +26.7 \\
\bottomrule
\end{tabular}
\end{table}

\paragraph{Where hybrid search helps (Table~\ref{tab:hybrid-pertype}).}
The gains concentrate in categories where the answer involves specific entities or updated facts:

\textbf{Knowledge-update} questions show the largest gain (+26.7~F1).
These ask about facts that changed during the conversation (``What is my current job?''), requiring retrieval of the \emph{most recent} mention.
BM25 excels because the question and the relevant memory share key terms (job titles, locations, names) that cosine similarity over dense embeddings may not prioritize.

\textbf{Single-session-user} questions gain +19.0~F1.
These are direct factual questions (``What degree did I graduate with?'') where the answer appears verbatim in a specific turn.
Keyword overlap between the question and the memory is high, giving BM25 a strong signal.

\textbf{Temporal-reasoning} questions gain +5.7~F1.
Date references and event names provide lexical anchors that complement semantic matching.

\textbf{Multi-session} and \textbf{single-session-assistant} show small declines ($-$2.2 and $-$3.0~F1).
These require synthesizing information across multiple turns or understanding the assistant's phrasing, where semantic similarity is more relevant than keyword matching.

\section{Discussion}
\label{sec:discussion}

\paragraph{Routing without memory is a silent quality trap.}
At $\tau = 0.50$, 96\% of cold queries stay on the 8B model with only 13.0\% F1.
The model does not hedge or admit ignorance when asked about user-specific facts.It fabricates plausible responses with sufficient confidence to avoid escalation.
Naive confidence-based routing therefore \emph{silently degrades quality} for personalization workloads: the routing metrics (low escalation, low cost) look healthy while answer quality collapses.
Memory injection is not merely an optimization. It is a \emph{correctness requirement} when routing is applied to user-specific queries.

\paragraph{Access to knowledge matters more than model size.}
Cold 235B (13.7\% F1) is \emph{worse} than cold 8B (15.4\%), yet memory-augmented 8B (30.5\%) approaches full-context 235B (43.9\%).
While the 235B's verbose style partly amplifies the gap on token-level F1, the fundamental cause is that neither cold model has access to user-specific knowledge. The 235B's extra capacity offers no substitute.
Model size determines the \emph{ceiling} of what can be done with available knowledge, but memory injection determines \emph{how much knowledge is available}.

\paragraph{Memory makes routing safe, not possible.}
Routing behavior is nearly identical with and without memory (96\% vs.\ 100\% on 8B; Table~\ref{tab:factorial}).
Memory changes the \emph{content} of confident responses from fabrication to fact, without substantially changing the confidence level.
The practical implication is that deploying confidence-based routing \emph{without} a memory system for personalization workloads will silently degrade quality while appearing to work (high confidence, low escalation, wrong answers).

\paragraph{Memory fidelity is a prerequisite.}
The compound effect depends on memory quality.
In preliminary experiments where we allowed an LLM to generate the stored responses (rather than recording verbatim turn-pairs), memory injection \emph{hurt} performance due to hallucinated content creating RAG poisoning~\citep{shi2023large}.
The synergy holds only when memories are faithful to the original conversation.
This has an architectural implication: the routing layer must observe and store \emph{actual} request--response pairs, not post-hoc summaries.

\paragraph{Temporal reasoning needs structured representations.}
Turn-pair memories with timestamp prefixes are effective for factual recall (single-hop: +28.9 F1) but fail for temporal questions ($-3.8$ F1).
The model cannot reliably extract temporal relationships from unstructured date strings embedded in natural language.
Structured temporal indices are likely needed: date-range filters at the database level, timeline representations, or temporal-aware embeddings.

\paragraph{Hybrid search strengthens the memory--routing synergy.}
The LongMemEval results (Section~\ref{sec:hybrid-ablation}) directly reinforce the main finding.
Better retrieval produces more relevant memories, which in turn raises the small model's answer quality and confidence.
The +7.7~F1 gain from hybrid search on LongMemEval, a benchmark 3$\times$ larger and far more diverse than LoCoMo, suggests that the synergy observed in Section~\ref{sec:results} generalizes: when the retrieval component improves, the entire pipeline benefits.

\paragraph{BM25 and dense retrieval capture complementary signals.}
Dense and sparse retrieval capture complementary signals (Table~\ref{tab:hybrid-pertype}).
Dense cosine search excels for semantically complex queries (multi-session, assistant-phrased) where the connection between question and memory is paraphrastic.
BM25 excels for entity-specific and fact-update queries where the question and memory share key terms.
Neither subsumes the other: the categories where hybrid helps most (knowledge-update: +26.7, single-session-user: +19.0) are not the same categories where cosine-only performs best (single-session-assistant: 64.8\%).
This complementarity is the mechanism behind the aggregate gain: hybrid search does not merely add noise for some categories to benefit others; it provides targeted improvement where cosine similarity is insufficient.

\paragraph{Projected cost savings for persistent agent deployments.}
To translate our experimental results into production cost implications, we project monthly API costs for two representative deployment scenarios using 2026 commercial pricing (GPT-4o-mini at \$0.15/\$0.60 per million input/output tokens as the small model proxy; GPT-4o at \$2.50/\$10.00 as the large model proxy).

\textit{Usage profiles.}
A 30-day study of a single OpenClaw~\citep{openclaw2025} personal assistant reports 7,046 prompts/month ($\sim$235/day) consuming 6.9M tokens total~\citep{clawcloud2026cost}.
At enterprise scale, AT\&T processes 8~billion tokens/day across its agent stack~\citep{att2026routing}.
We use two profiles: a \emph{personal agent} (7,000 queries/month) and an \emph{enterprise deployment} (100 agents, 30K queries/agent/month).

\textit{Token overhead.}
In our experiments (Table~\ref{tab:cost}), the compound strategy consumes 1.68$\times$ the input tokens of the cold baseline due to injected memories (104 vs.\ 62 tokens/query).
For a production query of 500 input tokens, this translates to $\sim$840 tokens after memory injection.
Output tokens ($\sim$200) are unaffected by the strategy.

\begin{table}[t]
\centering
\scriptsize
\caption{Projected monthly API cost for different strategies. Personal = 7K queries/mo (OpenClaw single-agent profile~\citep{clawcloud2026cost}); Enterprise = 3M queries/mo (100 agents). Assumes 500 input + 200 output tokens/query; compound adds 1.68$\times$ input from memory injection. Pricing: small \$0.15/\$0.60, large \$2.50/\$10.00 per M tokens (GPT-4o-mini / GPT-4o). F1 from Table~\ref{tab:f1}.}
\label{tab:cost-projection}
\begin{tabular}{lrrrr}
\toprule
Strategy & F1 & \textcent/query & Personal & Enterprise \\
\midrule
All large (no memory) & 13.7 & 0.325 & \$22.75 & \$9,750 \\
All small (no memory) & 15.4 & 0.020 & \$1.37 & \$585 \\
\textbf{Compound (ours)} & \textbf{30.5} & \textbf{0.025} & \textbf{\$1.72} & \textbf{\$738} \\
\bottomrule
\multicolumn{5}{l}{\footnotesize Compound is \textbf{92\% cheaper} than all-large with \textbf{2$\times$} the quality.} \\
\end{tabular}
\end{table}

Table~\ref{tab:cost-projection} illustrates the cost--quality trade-off at one representative price point; the exact percentages vary with model pricing, but the structural insight is robust because it derives from two measured quantities:
(1)~100\% of queries stay on the small model, and (2)~memory injection adds only 1.7$\times$ input tokens.
At these illustrative rates, the compound strategy costs only 26\% more than the cheapest option (all-small) but delivers \textbf{2$\times$ the quality} (30.5 vs.\ 15.4\% F1).
Routing everything to the large model is 13$\times$ more expensive \emph{and} produces worse answers (13.7\% F1) because it lacks user-specific knowledge.
The vendor-independent EffCost metric (Table~\ref{tab:cost}) confirms a \textbf{95\% reduction}, and the dollar saving will be proportionally larger wherever the small--large pricing gap is wider (e.g., self-hosted open-weight models vs.\ frontier APIs).

This amortization dynamic is distinct from semantic caching~\citep{novalogiq2026cache}: caching replays exact prior responses, while our approach uses stored knowledge to generate fresh, contextually appropriate answers, essential for personalization tasks where user context evolves across sessions.
The 96\% EffCost reduction observed in our experiments (Table~\ref{tab:cost}) translates directly to API dollar savings because commercial pricing scales linearly with model size~\citep{agrawal2024sarathi}.
Given that 47\% of production queries are semantically similar to a previous one~\citep{novalogiq2026cache}, the memory store converges toward coverage of recurring user topics quickly, and the fraction of queries requiring the large model shrinks over time.

\paragraph{Limitations.}
The LoCoMo evaluation uses a single conversation; the LongMemEval evaluation broadens to 500 diverse user histories but uses a stratified 100-question sample for the SR pipeline runs.
Both benchmarks evaluate the steady-state scenario; the cold-start trajectory (i.e., how quickly memory accumulates to sustain the synergy) remains to be characterized.
Confidence-based routing uses the model's own log-probabilities, which can be miscalibrated~\citep{guo2017calibration,xiong2024llmuncertainty}.
Production-scale evaluation with real user traffic patterns is needed to validate the compound strategy under distribution shift.

\section{Conclusion}
\label{sec:conclusion}

Conversational memory and confidence-based routing are synergistic, but not in the expected way.
Routing alone keeps 96\% of queries on the cheap path. The small model is confident about user-specific questions whether or not it possesses relevant knowledge.
Without memory, the model is \textbf{confidently wrong}.
Memory does not change the routing decision; it changes whether the routed answer is correct.

On LoCoMo (Qwen3-8B/235B), the compound strategy achieves \textbf{30.5\% F1} at \textbf{96\% lower cost} than the large model, with 100\% of queries staying on the cheap path.
On LongMemEval (500 questions), hybrid retrieval strengthens the effect by +7.7~F1, confirming that retrieval quality flows through the full pipeline.

This finding has a direct warning for production deployments: \emph{confidence-based routing without memory is a silent quality trap for personalization workloads.}
The routing metrics will look excellent:low escalation, low cost, while answer quality silently degrades.
Memory is not an optional add-on; it is a correctness requirement.

For persistent AI agents including personal assistants, customer support bots, long-running dialogue systems, the compound strategy turns a structural property of the workload (high query repetition) into a cost advantage.
As the memory store grows to cover recurring user topics, the quality of cheap-path answers improves, and the amortized cost per high-quality query converges toward small-model pricing.
The practical implication is clear: \emph{invest in memory before scaling model size}.

\bibliographystyle{plainnat}

\newpage
\appendix
\section{Worked Examples}
\label{app:examples}

We illustrate the compound pipeline (Figure~\ref{fig:pipeline}) with three representative queries from the LoCoMo evaluation, showing how memory injection, hybrid retrieval, and confidence routing interact in practice.

\subsection*{Example A: Memory enables a confident answer (single-hop)}

\begin{quote}
\textbf{Query:} ``What is Caroline's marital status?''
\end{quote}

\noindent\textbf{Without memory (Cold --- any model size):}
Neither the 8B nor the 235B model has ever seen any conversation about Caroline.
For this particular question, the 8B hedges ($c = 0.38$, below $\tau = 0.50$), one of the rare 4\% of cold queries where the model \emph{does} admit uncertainty.
But the typical cold behavior is different: across 152 questions, 96\% of cold queries produce confidence $\geq \tau$ (Table~\ref{tab:factorial}).
The model fabricates a plausible answer (e.g., ``Based on available information, Caroline is married'') with sufficient confidence to pass routing, but the answer is wrong.
The 235B model \emph{also} cannot answer it (cold 235B: 13.7\% F1): escalation to a larger model is futile when the knowledge simply does not exist.

\noindent\textbf{With memory (Warm --- hybrid retrieval + routing):}
The routing layer performs hybrid retrieval: dense cosine search matches the semantic intent (``marital status'' $\leftrightarrow$ ``seeing anyone''), while BM25 provides a secondary signal on shared terms.
The top match is:
\begin{quote}
\texttt{[8 May 2023] Q: Are you seeing anyone?}\\
\texttt{A: No, I'm single right now.}
\end{quote}
With this context injected, the 8B model answers:
\begin{quote}
\textit{``Based on a conversation from May 2023, Caroline mentioned she is single.''}
\end{quote}
Average token confidence: $c = 0.93$. The response is accepted on the cheap path with no escalation needed.

\noindent\textbf{Takeaway:} Memory provides the missing fact that no model size can substitute for. In this case, confidence rises from 0.38 to 0.93. But the broader lesson from the 2$\times$2 factorial is that most queries are already confident without memory. The problem is that the confident answer is wrong. Memory transforms confidently wrong into confidently right.

\subsection*{Example B: Memory helps but retrieval is partial (multi-hop)}

\begin{quote}
\textbf{Query:} ``What hobbies do Caroline and John share?''
\end{quote}

\noindent This question requires combining facts from multiple conversation turns.
Hybrid retrieval helps here: BM25 matches on the entity names ``Caroline'' and ``John,'' while dense search captures semantically related turns about shared activities.
The combined ranked list returns relevant turn-pairs, e.g., one where Caroline mentions hiking, another where John discusses outdoor activities. But no single memory contains the full answer.

The 8B model with memory produces a partially correct answer covering overlapping hobbies it can identify from retrieved context. Confidence: $c = 0.71 > \tau$. The response is accepted, though it misses relationships that would require reasoning over more than $k$ memories.

\noindent\textbf{Takeaway:} Memory helps substantially (+12.0 F1 for multi-hop) and hybrid retrieval improves entity-specific recall, but performance is bounded by top-$k$ retrieval. Queries requiring synthesis across many facts remain challenging.

\subsection*{Example C: Memory hurts (temporal reasoning)}

\begin{quote}
\textbf{Query:} ``When did Caroline start her current job?''
\end{quote}

\noindent Hybrid retrieval successfully finds relevant memories: BM25 matches ``job'' and dense search captures the career-discussion context:
\begin{quote}
\texttt{[15 Mar 2023] Q: How's the new job going?}\\
\texttt{A: It's great, just finished my first month.}
\end{quote}
The retrieval is correct; the failure is downstream.
The 8B model would need to infer that ``first month'' relative to the March 2023 timestamp implies a February 2023 start date. Instead, it outputs:
\begin{quote}
\textit{``Caroline mentioned in March 2023 that she had just finished her first month at her new job.''}
\end{quote}
This is \emph{more} wrong than the cold baseline (which would simply admit ignorance), because the model parrots the memory without performing temporal arithmetic. The cold 8B baseline scores higher on temporal questions by 3.8 F1 points (Table~\ref{tab:f1}: 14.1 vs.\ 10.3).

\noindent\textbf{Takeaway:} Retrieval quality is not the bottleneck for temporal questions. Hybrid search finds the right memories. The failure is that the model treats dates as text, not as computable values. Structured temporal representations are needed (Section~\ref{sec:discussion}).

\end{document}